\documentclass[letterpaper, 10 pt, conference]{ieeeconf}  

\IEEEoverridecommandlockouts                              
\usepackage{amsmath} 
\usepackage{amssymb}  

\usepackage{algorithm}
\usepackage{algpseudocode}
\usepackage{cite}

\newtheorem{definition}{Definition}
\newtheorem{assumption}{Assumption}
\newtheorem{theorem}{Theorem}
\newtheorem{corollary}{Corollary}

\usepackage{subcaption}
\usepackage{graphicx}

\title{\LARGE \bf
	DMPT-Swarm: A Directional Mass-Preserving Transform for Decentralized Swarm Navigation in Obstacle Environments
}

\author{Jacob Lee$^{1}$, Samuel Sutphin$^{1}$, Haroon Malik$^{2}$, Chang Liu$^{3}$, Pingping Zhu$^{1,2*}$ 
\thanks{This research was supported by the Defense Advanced Research Projects Agency (DARPA) -Grant \#000825 and the NASA Established Program to Stimulate Competitive Research (EPSCoR)-Grant \#80NSSC22M0027}
\thanks{$^{1}$Jacob Lee, Samuel Sutphin, and Pingping Zhu are with the Department of Biomedical and Electrical Engineering (CSEE), Marshall University, Huntington, WV 25755, USA (lee395@marshall.edu; sutphin54@marshall.edu; zhup@marshall.edu).}
\thanks{$^{2}$Haroon Malik is with the Department of Computer Science, Marshall University, Huntington, WV 25755, USA (malikh@marshall.edu)}
\thanks{$^{3}$Chang Liu are with the Department of Advanced Manufacturing and Robotics, College of Engineering, Peking University, Beijing 100871, China (changliucoe@pku.edu.cn).}
\thanks{All correspondences should be sent to Prof. Pingping Zhu.}}

\begin{document}

\maketitle
\thispagestyle{empty}
\pagestyle{empty}

\begin{abstract}

This paper presents a decentralized microscopic control framework for swarm navigation in obstacle environments, given a precomputed macroscopic trajectory represented by Gaussian mixture models (GMMs). To generate obstacle-aware reference locations without inter-agent communication, we first employ affine transport between multivariate normal (MVN) distributions under the Wasserstein-2 metric. We then propose a directional mass-preserving transform (DMPT) to incorporate obstacle effects by modifying radial transport while preserving directional structure. Based on the transformed references, a local artificial potential field (APF)-based controller is applied for navigation. Each agent operates independently, reducing communication and computational burden. Simulation results demonstrate improved efficiency and obstacle avoidance compared to conventional APF-based methods.

\end{abstract}

\section{INTRODUCTION}

Robotic swarm intelligence has emerged as a promising research area with applications in path planning, coverage, formation control, and navigation, ranging from autonomous drone swarms to large-scale fleets of mobile sensing systems. As the number of agents in a robotic swarm (RS) system increases, scalability and communication efficiency become critical challenges.

Existing swarm control methods are generally categorized into centralized and decentralized approaches. In centralized approaches, a global controller governs all agents by representing the swarm through a macroscopic state, typically modeled as spatial distributions such as Gaussian mixture models (GMMs) \cite{zhu2021adaptive, hu2024swarmprm, yang2024rover, gao2025swarmcvt, yang2025risk} or their variants, e.g., skew-normal mixture models (SNMMs) \cite{zhu2024novel}. These methods effectively address scalability at the macroscopic level. However, they require extensive data sharing among agents and the central unit, making them dependent on reliable, high-bandwidth communication, which is often impractical in real-world applications such as outdoor exploration or security-sensitive missions.
In contrast, decentralized approaches enable agents to make decisions individually or within local neighborhoods, thereby reducing communication requirements. However, they often suffer from increased computational complexity, as each agent must independently solve similar planning problems, leading to redundant computations and limited scalability in complex environments.

Most existing distribution-based swarm control frameworks adopt a two-layer hierarchical structure, consisting of macroscopic planning and microscopic control. At the macroscopic level, the swarm is represented by time-varying distributions, and planning is conducted in the probabilistic space to generate a desired trajectory of distributions. At the microscopic level, individual agents are controlled to follow this distribution. Although control inputs are applied at the agent level, such methods remain fundamentally centralized because the microscopic control laws depend on global distribution-level objectives and require knowledge of other agents’ states to minimize distribution discrepancies. This implicit coupling introduces significant communication overhead and limits scalability.

To address these challenges, this paper develops a fully decentralized microscopic control framework for swarm navigation in obstacle-populated environments, given a pre-computed macroscopic distribution trajectory from existing centralized planners \cite{zhu2021adaptive, hu2024swarmprm, yang2024rover, gao2025swarmcvt, yang2025risk, zhu2024novel}. The key idea is to decouple the global distribution-following objective into independent local control tasks without requiring inter-agent communication.

First, we employ an affine transform between multivariate normal (MVN) distributions, which corresponds to the optimal transport map under the Wasserstein-2 metric in obstacle-free environments, to generate reference locations for each agent. Next, we propose a novel directional mass-preserving transform (DMPT) to incorporate the effects of obstacles. The DMPT modifies the transport along radial directions while preserving directional structure, enabling efficient approximation of obstacle-aware skewed distributions. Finally, a local artificial potential field (APF)-based control law is designed to drive each agent toward its reference location while avoiding collisions. 

In the proposed framework, each agent computes its control input independently and locally, without requiring communication with other agents or a central unit. This significantly reduces computational complexity and communication burden compared to existing microscopic control strategies.
The main contributions of this paper are summarized as follows:

1. \textbf{Directional Mass-Preserving Transform:} We propose a novel transform for approximating obstacle-induced skew-normal distributions and generating obstacle-aware reference locations for individual agents. 

2. \textbf{Decentralized Microscopic Control:} We develop a fully decentralized control framework that eliminates inter-agent communication while maintaining distribution-level coordination.

3. \textbf{Validation and Performance Evaluation:} We validate the proposed DMPT and demonstrate the effectiveness of the control framework through numerical simulations.

\section{PROBLEM FORMULATION}

In this paper, we consider the problem of decentralized robotic swarm navigation in a large obstacle-populated region of interest (ROI) denoted by $\mathcal{W} \subset \mathbb{R}^2$. The robotic swarm (RS) or very-large-scale robotic (VLSR) system consists of $N\gg 1$ homogeneous cooperative mobile robots or agents, whose microscopic/individual time-varying position or state are denoted by $\mathbf{x}_n(t) \in \mathcal{W}$ for $n = 1, \ldots, N$ at time $t \in [t_0, t_f]$. To simplify the problem, the layout of obstacles is assumed to be static and known a priori, which can be represented by the set $\mathcal{B} = \bigcup_{m=1}^{M}\mathcal{B}_m \subset \mathcal{W}$, where $\mathcal{B}_m$ indicates the $m$th non-connected obstacle. Furthermore, we introduce the Hilbert map \cite{zhu2021adaptive} to represent the uncertainty of obstacles. Specifically, the probability that the location $\mathbf{x} \in \mathcal{W}$ is not occupied by obstacles is formulated as a Bernoulli-random-field (BRF), such that 
\begin{equation}
	Q_Y(\mathbf{x}) = Pr(Y = 0 |\mathbf{x} )
\end{equation}   
where $Y|\mathbf{x}$ is a Bernoulli random variable at the location $\mathbf{x}$, and $Y=0|\mathbf{x}$ indicates that the location is not occupied or obstacle-free.  

Similar to our previous papers \cite{zhu2021adaptive, hu2024swarmprm, yang2024rover, gao2025swarmcvt, yang2025risk}, let the macroscopic state of the swarm be described by a time-varying 2-D Gaussian mixture model (GMM) distribution, representing the time-varying spatial distribution of all agents, such that   
\begin{equation}
	\wp(\mathbf{x}, t_k) = \wp_k(\mathbf{x}) = \sum_{i=1}^{C_k} \omega_{k}^i  \phi_{\mathbf{X}}(\mathbf{x}|\boldsymbol{\mu}_{k}^{i}, \boldsymbol{\Sigma}_{k}^{i}) \in \mathcal{P(W)}
	\label{eq:GMM}
\end{equation}
where $\mathcal{P(W)}$ is the space of PDFs defined on $\mathcal{W}$. Here, we discretize the continuous time to the discrete or indexed time, such that the time variable $t  \in [0, t_f]$ is represented by the subscript $k$ for $k = 1, \ldots, T$, where $T = t_f / \Delta T$ and $\Delta T \ll t_f$. The term, $\wp_k(\mathbf{x})$, indicates the GMM at the $k$th time step $t_k = k \Delta T$, $C_k$ indicates the time-varying number of Gaussian components (GCs), and $\phi_{\mathbf{X}}(\mathbf{x}|\boldsymbol{\mu}_{k}^{i}, \boldsymbol{\Sigma}_{k}^{i})$ indicates the $i$th Gaussian component/MVN distribution with mean $\boldsymbol{\mu}_{k}^{i}$ and covariance $\boldsymbol{\Sigma}_{k}^{i}$, for $i=1,\ldots,C_k$, associated with the component weight $\omega_{k}^{i} \in (0, 1)$, such that $\sum_{i=1}^{C_k} \omega_{k}^{i} = 1$, where the superscript ``$i$" indicates the index of GCs. 

%

The goal of this paper is to develop the decentralized microscopic control, instead of the centralized macroscopic path planning methods, which have been developed in our previous work \cite{zhu2021adaptive, hu2024swarmprm, yang2024rover, gao2025swarmcvt, yang2025risk}. Let $\mathcal{X}_k = \{\mathbf{x}_{n,k}\}_{n=1}^N$ denote the set of agent states at time step $k$, e.g., $\mathbf{x}_{n,k} = \mathbf{x}_n(k \Delta t)$. Given the trajectory of the GMMs, denoted by $\mathcal{T}_{1:T} = [ \wp_1, \ldots, \wp_T ]$ and the obstacle set $\mathcal{B}$, the decentralized control law is designed such that the empirical distribution of $\mathcal{X}_k$ approximates the prescribed GMM $\wp_k$, while all agents remain in the obstacle-free region. Specifically, we need to design a continuous control law 
\begin{equation}
	\dot{\mathbf{x}}_{n}(t) = C(\mathbf{x}_{n}(t) | \mathcal{T}_{1:T}) 	 
	\label{eq:decentralizedControlLaw}
\end{equation}     
to minimize the following instant cost function at each discrete time step,
\begin{equation}
	J(\mathcal{X}_k) = D(\wp_k, \hat{\wp}_k) + \lambda \sum_{n=1}^N \left[1 - Q_Y(\mathbf{x}_{n,k})\right]
	\label{eq:costFuncTotal}
\end{equation}
where $D(\wp_k, \hat{\wp}_k)$ indicates the measure of dissimilarity between the reference distribution $\wp_k$ and the empirical distribution of all agents at the time $t = k\Delta t$, and $\lambda$ is user-defined weighted parameter. 


It is noteworthy that we aim to implement the decentralized control in (\ref{eq:decentralizedControlLaw}) without communication, where no data sharing occurs among agents during the decentralized swarm navigation.


\section{APPROXIMATION OF BRF-SN DISTRIBUTIONS}
This section reformulates the d-Dimensional BRF-SN distribution in the direction-radial coordinate, and approximate it by the 1-D skew distributions along a varying direction specified by the point.  

\subsection{Multivariate Normal Distributions in Directional Radial Coordinates}
Multivariate normal (MVN) distributions, also called Gaussian distributions, are widely used in control and robotics because of their analytical tractability and their ability to represent anisotropic uncertainty. Let $\mathbf{X} \in \mathbb{R}^d$ be an MVN random vector with mean $\boldsymbol{\mu} \in \mathbb{R}^d$ and covariance matrix 
$\boldsymbol{\Sigma} \in \mathbb{R}^{d \times d}$, denoted by $\mathbf{X} \sim \mathcal{N}(\boldsymbol{\mu}, \boldsymbol{\Sigma})$. Its probability density function (PDF), $\phi_{\mathbf{X}}(\mathbf{x}|\boldsymbol{\mu}, \boldsymbol{\Sigma})$, is given by
\begin{equation}
	\phi_{\mathbf{X}}(\mathbf{x})=\frac{1}{\sqrt{(2\pi)^d|\boldsymbol{\Sigma}|}}\exp{\left(-\frac{(\mathbf{x}-\boldsymbol{\mu})^T \boldsymbol{\Sigma}^{-1}(\mathbf{x}-\boldsymbol{\mu})}{2}\right)}
	\label{eq:MVN}
\end{equation}
where $\mathbf{x} \in \mathbb{R}^d$, and the parameters of $\boldsymbol{\mu}$ and $\boldsymbol{\Sigma}$ are suppressed from $\phi_{\mathbf{X}}(\mathbf{x}|\boldsymbol{\mu}, \boldsymbol{\Sigma})$ as shorthand.

Let $\mathbf{u} \in \mathbb{U}^{d-1}_+ \subset \mathbb{R}^d$ indicate a unit vector, where $\mathbb{U}^{d-1}_+ = \{\mathbf{u} | \mathbf{u} \in  \mathbb{R}^d,  \Vert \mathbf{u}\Vert = 1,  \mathbf{u}^T \mathbf{e}_1 \geq 0\}$ indicates the positive hemisphere of $\mathbb{R}^d$ where $\mathbf{e}_1 \in \mathbb{R}^d$ is the unit vector along the first coordinate axis. Let $s \in \mathbb{R}$ indicate the  signed radial. Then, given $\boldsymbol{\mu}$, the vector $\mathbf{x}\in \mathbb{R}^d$ in the Cartesian coordinate and $(s,\mathbf{u})$ in the directional-radial coordinate can be converted via the following transforms, 
\begin{align}
	\mathbf{x} &= T_C(s, \mathbf{u}; \boldsymbol{\mu}) = s \mathbf{u} + \boldsymbol{\mu} \\
	s &= T_S(\mathbf{x}; \boldsymbol{\mu}) = \alpha(\mathbf{x}; \boldsymbol{\mu})  \Vert \mathbf{x} - \boldsymbol{\mu} \Vert \\
	\mathbf{u} &= T_U(\mathbf{x}; \boldsymbol{\mu}) =  \alpha(\mathbf{x}; \boldsymbol{\mu}) \frac{\mathbf{x} - \boldsymbol{\mu}}{\Vert \mathbf{x} - \boldsymbol{\mu} \Vert} 
\end{align}
where $\alpha(\mathbf{x}; \boldsymbol{\mu})$ is defined by
\begin{equation}
	\alpha(\mathbf{x}; \boldsymbol{\mu}) = \begin{cases}
		1 & \text{ if } (\mathbf{x} - \boldsymbol{\mu})^T \mathbf{e}_1 \geq 0 \\
		-1 & \text{ if } (\mathbf{x} - \boldsymbol{\mu})^T \mathbf{e}_1 < 0
	\end{cases}
\end{equation} 
The representation is unique for all $\mathbf{x} \neq \boldsymbol{\mu}$. The point $\mathbf{x} = \boldsymbol{\mu}$ is singular and may be assigned arbitrarily since it has measure zero under a continuous MVN distribution.

Furthermore, let $\mathbf{U} \in \mathbb{U}_+^{d-1}$ and $S \in \mathbb{R}$ denote the unit random vector and the scalar random variable, respectively. Then, the random vector $\mathbf{X}$ can be expressed by
\begin{equation}
	\mathbf{X} (S, \mathbf{U}) = T_C (S, \mathbf{U} ; \boldsymbol{\mu}) = S \mathbf{U} + \boldsymbol{\mu} 
\end{equation}  
Specifically, for the 2-dimensional case $d = 2$, the unit vector would be $\mathbf{U}(\Theta) = [\cos (\Theta), \sin (\Theta) ]^T$, where $\Theta$ indicates the angle random variable $\Theta \in [0, \pi] \in \mathbb{R}$.

Therefore, the joint PDF of $(S, \mathbf{U})$ given $\boldsymbol{\mu}$ and $\boldsymbol{\Sigma}$ is expressed by
\begin{equation}
	f_{S,\mathbf{U}}(s,\mathbf{u}) = \det (J_{T_C})   \phi_{\mathbf{X}}(\mathbf{x}) = \vert s \vert ^{d-1}  \phi_{\mathbf{X}}(s \mathbf{u} + \boldsymbol{\mu})
	\label{eq:f_S_U}
\end{equation}
where $\det (J_{T_C}) = \vert s \vert ^{d-1}$ indicates the determinant of the Jacobian matrix $J_{T_C}$ with respect to the transform $T_C$. In addition, we have the volume element relationship
\begin{equation}
	d\mathbf{x} =  \vert s \vert ^{d-1}  ds d \sigma(\mathbf{u}) 
	\label{eq:delta_transform}
\end{equation} 
 
It is worth noting that $f_{S,\mathbf{U}}(s,\mathbf{u})$ specifies a directional-radial coordinate PDF for the MVN random vector $\mathbf{X}$ in the Cartesian coordinates, which is constructed based on the geometric decomposition. 


Furthermore, by integrating out $s\in \mathbb{R}$ in (\ref{eq:f_S_U}), the marginal density of $\mathbf{U}$ can be obtained, such that
\begin{align}
	f_{\mathbf{U}} (\mathbf{u}) &= \int_{-\infty}^{\infty}  \vert s \vert ^{d-1}  \phi_{\mathbf{X}}(s \mathbf{u} + \boldsymbol{\mu}) ds \nonumber\\
	&= \frac{\Gamma(d/2)}{\pi^{d/2} \vert \boldsymbol{\Sigma}\vert^{1/2}} \left(\mathbf{u}^T \boldsymbol{\Sigma}^{-1}\mathbf{u}\right)^{-d/2} 
	\label{eq:f_U}
\end{align}  
It is noteworthy that $f_{\mathbf{U}} (\mathbf{u})$ is uniform on $\mathbf{U}^{d-1} $ for $\boldsymbol{\Sigma} = \lambda \mathbf{I}_d$, where $\mathbf{I}_d$ indicates the $d \times d$ identity matrix. Specifically, for 2-D case by substituting $d = 2$, we have
\begin{equation}
	f_{\mathbf{U}}(\mathbf{u}) = \frac{1}{\pi \vert \boldsymbol{\Sigma}\vert^{1/2}} \frac{1}{\mathbf{u}^T \boldsymbol{\Sigma}^{-1}\mathbf{u}}
\end{equation}
and for $\boldsymbol{\Sigma} = \lambda \mathbf{I}_2$ we have $f_{\mathbf{U}}(\mathbf{u}) = 1/\pi$, which represents the special case that $\Theta$ is uniform distributed on $[0, \pi]$.   

Finally, according to the Bayesian rule, we can obtain the conditional PDF, $f_{S|U}(s |\mathbf{u})$, such that
\begin{align}
	f_{S|\mathbf{U}}(s|\mathbf{u}) &= \frac{f_{S,\mathbf{U}}(s, \mathbf{u})}{f_{\mathbf{U}}(\mathbf{u})}  \\
	&=   \frac{\left(\mathbf{u}^T \boldsymbol{\Sigma}^{-1}\mathbf{u}\right)^{d/2}  \vert s \vert ^{d-1}}{2^{d/2}\Gamma(d/2)}        \exp(-\frac{s^2 \mathbf{u}^T \boldsymbol{\Sigma}^{-1} \mathbf{u}}{2}) \nonumber 
	\label{eq:conditionalPDF_S}
\end{align}
which is a symmetric double-sided Chi-type distribution.

Specifically, for 2D case, substituting $d=2$, we have
\begin{align}
	f_{S|\mathbf{U}}(s|\mathbf{u}) 
	&= \frac{\vert s \vert}{ 2 \sigma_R^2(\mathbf{u})}  \exp(-\frac{s^2}{2 \sigma_R^2(\mathbf{u})}) 
\end{align}
It shows that for the 2-D case, the conditional density $f_{S|\mathbf{U}=\mathbf{u}}(s)$ corresponds to a symmetric double-sided Rayleigh-type distribution, such that $[ \left. \vert S  \vert \right\vert \mathbf{U}=\mathbf{u} ]  \sim Rayleigh(\sigma_R)$, where $\sigma_R(\mathbf{u}) = (\mathbf{u}^T \boldsymbol{\Sigma}^{-1} \mathbf{u})^{-1/2}$ indicates the Rayleigh parameter. The corresponding cumulative distribution function (CDF) of the double-sided Rayleigh distribution admits a closed-form exponential expression, 
\begin{equation}
	F_{S|\mathbf{U} }(s | \mathbf{u}) = \begin{cases}
		\frac{1}{2} \exp (-\frac{s^2}{2\sigma_R^2}), & \text{for } s<0 \\
		1 - \frac{1}{2} \exp (-\frac{s^2}{2\sigma_R^2}), & \text{for } s\geq0
	\end{cases}
	\label{eq:doubleRayleighCDF}
\end{equation}

\subsection{Multivariate Skew-Normal Distributions}
Although MVN distributions are commonly used to model the spatial distribution of agents in VLSR systems, their unbounded support is inconsistent with environments that contain obstacles. Thus, we have proposed the Bernoulli-random-field based skew-normal (BRF-SN) distribution in \cite{zhu2024novel} to reflect the repulsive force from obstacles to the MVN distributions, which is defined in the following definition. 

\begin{definition}[BRF-SN \cite{zhu2024novel}] 
	\label{def:BRF-SND}
	Consider a continuous random vector,  $\mathbf{X} \in \mathbb{R}^d$ and a Bernoulli random variable $Y \in \{0, 1\}$, where the random vector $\mathbf{X}$ is multivariate normal  distributed, such that  $\mathbf{X} \sim \mathcal{N}(\boldsymbol{\mu},\boldsymbol{\Sigma})$. The random variable,  $\mathbf{X}^* = \left[ \mathbf{X} \vert Y = 0 \right]$, has a Bernoulli-random-field based skew-normal (BRF-SN) distribution denoted by $\mathbf{X}^* \sim BRFSN(\boldsymbol{\mu}, \boldsymbol{\Sigma}, Q_Y)$, if its PDF is given by
	\begin{equation}
		f^*_{\mathbf{X}}(\mathbf{x}; \boldsymbol{\mu}, \boldsymbol{\Sigma}) = \frac{\phi_{\mathbf{X}} (\mathbf{x} \vert \boldsymbol{\mu}, \boldsymbol{\Sigma}) Q_Y(\mathbf{x})}{\mathbb{E}_{\mathbf{X}\sim \phi_{\mathbf{X}}}[Q_Y(\mathbf{X})]}
		\label{eq: PDF_BRFSN}
	\end{equation} 
	where $\phi_{\mathbf{X}} (\mathbf{x} \vert \boldsymbol{\mu}, \boldsymbol{\Sigma})$ is the PDF of the multivariate normal random vector $\mathbf{X}$, and $Q_Y(\mathbf{x}) \triangleq Pr(Y(\mathbf{X}) = 0 \vert \mathbf{X} = \mathbf{x})$. 
\end{definition}

In \textit{Definition \ref{def:BRF-SND}}, the skewing function $Q_Y(\mathbf{x})$ defines a Bernoulli-random field (BRF), which indicates the probability that the location $\mathbf{x} \in \mathbb{R}^d$ is not occupied by obstacles. The skewing function can be obtained or incrementally updated through many approaches, e.g., the Hilbert occupancy map. The readers are referred to \cite{ramos2016hilbert, morelli2019integrated, zhu2019scalable}. In the rest of this paper, we use the superscript ``*" to indicate skewed random variables and their corresponding PDFs and CDFs.

\subsection{Directional-Wise Approximation of BRF-SN Density}
\label{sec:ApproxBRFSN}
By substituting (\ref{eq:f_S_U}) and (\ref{eq:delta_transform}) into (\ref{eq: PDF_BRFSN}), the BRF-SN $f^*_{\mathbf{X}}(\mathbf{x})$ defined in \textit{Definition} \ref{def:BRF-SND} can be re-expressed by

\begin{align}
	f^*_{\mathbf{X}}(\mathbf{x}) =& \frac{ \vert s \vert^{1-d}	f_{S,\mathbf{U}}(s,\mathbf{u})   Q_Y(\mathbf{x} )}{\mathbb{E}_{\mathbf{X}\sim \phi_{\mathbf{X}}}[Q_Y(\mathbf{X})]} \nonumber\\
	=& \vert s \vert^{1-d} f^*_{S,\mathbf{U}}(s,\mathbf{u})
\end{align}
Here, $f^*_{S,\mathbf{U}}(s,\mathbf{u})$ indicates the directional-radial form of the BRF-SN PDF, such that
\begin{align}
	&f^*_{S,\mathbf{U}}(s,\mathbf{u}) \triangleq \frac{	f_{S,\mathbf{U}}(s,\mathbf{u})   Q_Y(\mathbf{x} )}{\mathbb{E}_{\mathbf{X}\sim \phi_{\mathbf{X}}}[Q_Y(\mathbf{X})]} \nonumber\\ 
	=& \frac{f_{S,\mathbf{U}}(s,\mathbf{u})   Q_Y(s, \mathbf{u} )}{\int_{\mathbf{x}^{\prime}\in \mathbb{R}^d} \phi_{\mathbf{X}} (\mathbf{x}^{\prime} \vert \boldsymbol{\mu}, \boldsymbol{\Sigma}) Q_Y(\mathbf{x}^{\prime}) d\mathbf{x}^{\prime}} \nonumber\\
	=& \frac{ f_{\mathbf{U}}(\mathbf{u}) f_{S|\mathbf{U}}(s, \mathbf{u})  Q_Y(s, \mathbf{u})}{\int_{\mathbf{u}^{\prime} \in \mathbb{U}^{d-1}_+} f_{\mathbf{U}}(\mathbf{u}^{\prime}) \mathbb{E}_{S|\mathbf{U}}[Q_Y(S, \mathbf{u}^{\prime} )] d \sigma(\mathbf{u}^{\prime}) } \nonumber\\
	=& \frac{	f_{S,\mathbf{U}}(s,\mathbf{u})   Q_Y(s,\mathbf{u} )}{\mathbb{E}_{S,\mathbf{U}}[Q_Y(S, \mathbf{U})]}	
	\label{eq:f_SU*}
\end{align}
where the term, $Q_Y(s, \mathbf{u})$, indicate the skew function in the directional-radial coordinate, such that
\begin{equation}
	Q_Y(s, \mathbf{u}) = Q_Y(s, \mathbf{u} ;  \boldsymbol{\mu}) = Q_Y(s \mathbf{u} + \boldsymbol{\mu}) = Q_Y(\mathbf{x})
\end{equation}
and 
\begin{equation}
	\mathbb{E}_{S|\mathbf{U}=\mathbf{u}}[Q_Y(S, \mathbf{u})] = \int_{s^{\prime} \in \mathbb{R}}  f_{S|\mathbf{U}}(s^{\prime} | \mathbf{u})  Q_Y(s^{\prime},  \mathbf{u})  ds^{\prime}
\end{equation}

Because of the arbitrariness of the skew function $Q_Y(\mathbf{x})$, there exists no closed form for the denominator in (\ref{eq: PDF_BRFSN}) and (\ref{eq:f_SU*}). The calculation of the denominator depends on the values of $Q_Y(\mathbf{x})$ on the whole space $\mathbb{R}^d$, which is time-consuming even when approximated within the workspace $\mathcal{W}\subset \mathbb{R}^d$. To address this issue, we introduce the following assumption.

\begin{assumption}
\label{ass:directionalSkewFunc}
 The agent or particle $\mathbf{x}_0 \in \mathbb{R}^d$ is only allowed to move along the corresponding direction $\mathbf{u}_0 = T_U(\mathbf{x}_0)$ when it is affected by the repulsive force from the obstacles. It means for the agent $\mathbf{x}_0$, the skew function $Q_Y(\mathbf{x})$ can be approximated by
\begin{equation}
	\tilde{Q}_Y(\mathbf{x}, \mathbf{u}_0) = Q_Y(\mathbf{x}) \boldsymbol{1}_{\mathbb{L}(\mathbf{u}_0)}(\mathbf{x})
	\label{eq:tildeQ}
\end{equation}
Here, $\boldsymbol{1}_{\mathbb{L}(\mathbf{u})}(\mathbf{x})$ is an indicator function with the set $\mathbb{L}(\mathbf{u}) = \mathbb{L}(\mathbf{u};\boldsymbol{\mu}) = \{\mathbf{x} | \mathbf{x} = s \mathbf{u} + \boldsymbol{\mu} \text{ for } s \in \mathbb{R}  \}$. 
\end{assumption}
 
This approximation assumes that the variation of $Q_Y(\mathbf{x})$ in directions orthogonal to $\mathbf{u}_0$ is negligible locally. This assumption reflects a local control policy in which the agent reacts primarily to obstacles along a dominant direction $\mathbf{u}_0$, thereby reducing directional uncertainty for computational tractability. 

Then, by replacing $Q_Y(\mathbf{x})$ in (\ref{eq: PDF_BRFSN}) with  $\tilde{Q}_Y(\mathbf{x}, \mathbf{u})$ provided in (\ref{eq:tildeQ}), first, we can obtain the approximate BRF-SN PDF, 
such that
\begin{align}
	\tilde{f}^*_{\mathbf{X}}(\mathbf{x}; \boldsymbol{\mu}, \boldsymbol{\Sigma}) &= \frac{\phi_{\mathbf{X}} (\mathbf{x} \vert \boldsymbol{\mu}, \boldsymbol{\Sigma}) \tilde{Q}_Y(\mathbf{x}, \mathbf{u})}{\mathbb{E}_{\mathbf{X}\sim \phi_{\mathbf{X}}}[\tilde{Q}_Y(\mathbf{X}, \mathbf{u})]} \nonumber\\
	&= \frac{\phi_{\mathbf{X}} (\mathbf{x} \vert \boldsymbol{\mu}, \boldsymbol{\Sigma}) Q_Y(\mathbf{x}) \boldsymbol{1}_{\mathbb{L}(\mathbf{u})}(\mathbf{x})}{\mathbb{E}_{\mathbf{X}\sim \phi_{\mathbf{X}}}[Q_Y(\mathbf{X})\boldsymbol{1}_{\mathbb{L}(\mathbf{u})}(\mathbf{X})]}\nonumber\\
	&= \frac{\phi_S(s|0,\sigma^2_R(\mathbf{u})) Q_Y(s \mathbf{u} + \boldsymbol{\mu} ) }{\mathbb{E}_{S\sim \phi_S}[Q_Y(S \mathbf{u} + \boldsymbol{\mu})]} \nonumber\\
	&\triangleq f_{S,\mathbf{U}}^*(s, \mathbf{u}; \boldsymbol{\mu}, \boldsymbol{\Sigma})
	\label{eq: PDF_BRFSN_Approx}
\end{align}
where $\phi_S(s|0,\sigma^2_R) = \phi_{\mathbf{X}} (s \mathbf{u} + \boldsymbol{\mu} \vert \boldsymbol{\mu}, \boldsymbol{\Sigma})$ indicates a 1-D zero-mean Gaussian/normal distribution with the variance of $\sigma_R^2 = (\mathbf{u}^T \boldsymbol{\Sigma}^{-1} \mathbf{u})^{-1}$. It shows that under \textit{Assumption-1}, the $d-D$ BRF-SN can be approximated by the 1-D BRF-SN along the direction $\mathbf{u} = T_U(\mathbf{x})$. 

Second, similarly, by replacing $Q_Y(\mathbf{x})$ in (\ref{eq:f_SU*}) with  $\tilde{Q}_Y(\mathbf{x}, \mathbf{u})$, we can have  
\begin{align}
	\tilde{f}^*_{S,\mathbf{U}}(s,\mathbf{u}) =& \frac{	f_{S,\mathbf{U}}(s,\mathbf{u})   Q_Y(s,\mathbf{u} )\boldsymbol{1}_{\mathbb{L}(\mathbf{u})}(s\mathbf{u} + \boldsymbol{\mu})}{\mathbb{E}_{S,\mathbf{U}}[Q_Y(S, \mathbf{U}) \boldsymbol{1}_{\mathbb{L}(\mathbf{u})}(S \mathbf{U} + \boldsymbol{\mu}) ]}	 \nonumber\\ 
	=& \frac{ f_{S,\mathbf{U}}(s,\mathbf{u})  Q_Y(s, \mathbf{u})}{ f_{\mathbf{U}}(\mathbf{u}) \mathbb{E}_{S|\mathbf{U}}[Q_Y(S, \mathbf{u} )]  } \nonumber\\
	=& \frac{  f_{S|\mathbf{U}}(s | \mathbf{u})  Q_Y(s, \mathbf{u})}{  \mathbb{E}_{S|\mathbf{U}}[Q_Y(S, \mathbf{u} )]  } \nonumber\\	
	\triangleq& f^*_{S|\mathbf{U} }(s | \mathbf{u})
	\label{eq:PDF_BRFSN_FixU_Approx}
\end{align}
Under \textit{Assumption \ref{ass:directionalSkewFunc}}, the joint distribution degenerates into a conditional distribution along the fixed direction $\mathbf{u}=T_U(\mathbf{x})$. Here, we define a new skew distribution $f^*_{S|\mathbf{U}}(s | \mathbf{u})$ based on the conditional distribution $f_{S|\mathbf{U}}(s |\mathbf{u})$. Since $f_{S|\mathbf{U}}(s|\mathbf{u})$ is a symmetric double-sided Chi distribution (Rayleigh distribution for 2-D case), $f^*_{S|\mathbf{U}}(s| \mathbf{u})$ specifies a Skew-Chi distribution or Skew-Rayleigh distribution for 2-D case.

%

\section{DIRECTIONAL-WISE TRANSFORM FOR APPROXIMATED BRF-SN}
In this section, two directional mass-preserved transforms (DMPTs) are proposed to convert the $d$-D MVN distribution to approximate the corresponding BRF-SN distributions. 

\subsection{1-D Distribution Representation of Points}
For the 1-D case, we can define the following distribution to represent a group of agents/points along a direction. 
\begin{definition}
	\label{def:distributionRepresentation}
	Let $\mathcal{S} = \{s_n\}_{n=1}^N$ indicates a set of $N$ points $s_n \in \mathbb{R}$ for $n = 1,\ldots, N$. Without loss of generality, assume that the elements in $\mathcal{S}$ are all distinct, such that $s_1 < s_2 < \ldots <  s_N$. A continuous probabilistic distribution of the random variable, $S \in \mathbb{R}$, is said to be able to represent the distribution of the set, $\mathcal{S}$, if for any point $s_n \in \mathcal{S}$, its cumulative distribution function (CDF), $F_S(s)$, satisfies,
	\begin{align}
		F_S(s_n) & > \frac{\sum_{s_i \in \mathcal{S}} \boldsymbol{1}_{\mathbf{S}_L(s_n)}(s_i) }{N} , \; \forall s_n \in \mathcal{S}\\
		1 - F_S(s_n) & > \frac{\sum_{s_i \in \mathcal{S}} \boldsymbol{1}_{\mathbf{S}_R(s_n)}(s_i) }{N}, \; \forall s_n \in \mathcal{S}
	\end{align} 
	where $\mathcal{S}_L(s_n) = \{s_i | s_i \in \mathcal{S}, s_i < s_n\}$ and $\mathcal{S}_R(s_n) = \{s_i | s_i \in \mathcal{S}, s_i > s_n\}$. 
\end{definition}

Here, $\sum_{s_i \in \mathcal{S}} \boldsymbol{1}_{\mathbf{S}_L(s_n)}(s_i)$ and $\sum_{s_i \in \mathcal{S}} \boldsymbol{1}_{\mathbf{S}_R(s_n)}(s_i)$ count the numbers of points lying to the left and right of $s_n$, respectively. Obviously, based on \textit{Definition \ref{def:distributionRepresentation}} there are many probabilistic distributions that can be applied to represent the same point set. Furthermore, we have the following theorem.
\begin{theorem}
	\label{theorem:TwoDistributionsRepresentation}
	Given a set of $N$ distinct points, $\mathcal{S} = \{s_n\}_{n=1}^N$, where $s_n \in \mathbb{R}$, there are two different continuous probabilistic distributions that can represent $\mathcal{S}$ associated with the two CDFs, $F_1(s)$ and $F_2(s)$, respectively. Then, we have the following inequality,
	\begin{equation}
		\vert F_1(s_n) - F_2(s_n) \vert < \frac{1}{N}, \; \forall s_n \in \mathcal{S} 
	\end{equation}    
\end{theorem}

The proof of \textit{Theorem \ref{theorem:TwoDistributionsRepresentation}} is provided in Appendix \ref{proof:Theorem-1}. In addition, we have the following corollary:



\begin{corollary}
	\label{cor:radialTrans}
	Consider a set of $N$ ordered points $S=\{s_n\}_{n=1}^N$, where
	$s_n\in\mathbb{R}$ and $s_n<s_m$ if $n<m$, whose distribution can be
	represented by a continuous probabilistic distribution with CDF $F(s)$.
	Let $T_{OP}:\mathbb{R}\to\mathbb{R}$ be an order-preserving transform, and
	define $s_n' = T_{OP}(s_n)$, for $n=1,\dots,N$, such that 
	\begin{equation}
			s_n^{\prime} < s_m^{\prime} \text{ if } s_n < s_m
	\end{equation}  
	Let $S'=\{s_n'\}_{n=1}^N$, and suppose that a continuous probabilistic
	distribution with CDF $F_{OP}(s')$ can represent $S'$. Then,
	\begin{equation}
		\vert F(s_n) - F_{OP}(s_n^\prime) \vert < \frac{1}{N} \text{ for } n = 1,\ldots, N
		\label{eq:PDF_diff}
	\end{equation} 
\end{corollary}

\textit{Corollary \ref{cor:radialTrans}}  follows directly from \textit{Theorem \ref{theorem:TwoDistributionsRepresentation}} and the order-preserving property of the transform.


\subsection{Directional Mass-Preserved Transform for BRF-SN}

In \cite{zhu2024novel}, the BRF-SN distribution is introduced to model the repulsive effect of obstacles on agents, where the nominal agent distribution is assumed to follow an MVN distribution in the absence of obstacles. Due to obstacle-induced forces, the global spatial structure of the MVN distribution is generally distorted. However, it is reasonable to assume that along any fixed direction, the relative ordering of agents is approximately preserved. This is because the repulsive effects of obstacles act consistently along a given direction, inducing a directional displacement rather than arbitrary reordering. Therefore, when the d-dimensional BRF-SN distribution is projected onto a one-dimensional subspace along a fixed direction, the resulting transformation can be reasonably approximated as order-preserving. 

In light of \textit{Corollary \ref{cor:radialTrans}}, we can construct the following directional mass preserving transform (DMPT), $T_{DMPT}$, to implement the transport points from the MVN distribution to approximate BRF-SN distributions. 

\begin{definition}
	Consider a set of $N$ points in $d$-D space, $\mathcal{X} = \{ \mathbf{x}_n \}_{n=1}^N  \subset \mathbb{R}^d$, which can be represented by a MVN distribution with the random vector $\mathbf{X}$ and the PDF, $\phi_{\mathbf{X}}(\mathbf{x}|\boldsymbol{\mu}, \boldsymbol{\Sigma})$. Due to the obstacles specified by the skew function $Q_Y(\mathbf{x})$, the points set is deformed to a new set of points, $\mathcal{X}^{\prime} = \{ \mathbf{x}^{\prime}_n \}_{n=1}^N \subset \mathbb{R}^d$, which can also be represented by a continuous probabilistic distribution with the random vector $\mathbf{X}^*$,  by the following transform,
	\begin{align}
		T_{DMPT} : \mathbb{R}^d &\to \mathbb{R}^d \\
		\mathbf{x} &\mapsto \mathbf{x}^{\prime} = T_{DMPT }(\mathbf{x} | \phi_{\mathbf{X}}, Q_Y) = s^{\prime} \mathbf{u}^\prime + \boldsymbol{\mu}  \nonumber
		\label{eq:DMPT}
	\end{align}
	where 
	\begin{align}
		\mathbf{u}^\prime &= T_U(\mathbf{x}^\prime) = T_U(\mathbf{x}) \\
		s^{\prime} &= T_S(\mathbf{x}^{\prime}) = {F^*_S}^{-1}(F_S (s)) = {F^*_S}^{-1}(F_S (T_S(\mathbf{x}))) \label{eq:inverseCalculation}
	\end{align}
	Here, $F_S (s)$ and $F^*(s)$ are a pair CDFs defined on $\mathbb{R}$ for the random variable $S = T_S(\mathbf{X})$ and $S^* = T_S(\mathbf{X}^*)$.   
\end{definition}   

\textit{Remarks:} The transform above is proposed inspired by (\ref{eq:PDF_diff}) in \textit{Corollary  \ref{cor:radialTrans}}. The CDF $F(s_n)$ monotonically approaches to $F_{OP}(s^{\prime}_n)$ when the number of agents is increasing, $N \to \infty$. Even if $N$ is small, the absolute error of CDFs in (\ref{eq:PDF_diff}) is also bounded. Therefore the proposed transform corresponds to a one-dimensional optimal transport map under the monotonicity constraint.   

Based on Section \ref{sec:ApproxBRFSN}, furthermore, we propose two transforms according to (\ref{eq: PDF_BRFSN_Approx}) and (\ref{eq:PDF_BRFSN_FixU_Approx}).  First, for a given direction $\mathbf{u} = \boldsymbol{\xi}$, let $F_S(s) = \Phi_S(s|0, \sigma_R(\boldsymbol{\xi}))$ denote the CDF of the MVN distribution and $F^*_S(s) = F_{S,\mathbf{U}}^*(s, \boldsymbol{\xi}; \boldsymbol{\mu}, \boldsymbol{\Sigma})$ denote the CDF of $f_{S,\mathbf{U}}^*(s, \boldsymbol{\xi}; \boldsymbol{\mu}, \boldsymbol{\Sigma})$ in (\ref{eq: PDF_BRFSN_Approx}). Due to this pair of MVN and BRF-SN distributions, we call the transform SN-based DMPT (DMPT-SN), denoted by $T_{DR}^{SN}$.  Therefore, given $\mathbf{X} \sim \mathcal{N}(\boldsymbol{\mu}, \boldsymbol{\Sigma})$,  the transformed V.R. $\mathbf{X}^{\prime} = T_{DMPT}^{SN}(\mathbf{X})$  admits the directional-radial density and Cartesian PDF, such that
\begin{align}
	f_{S^{\prime}, \mathbf{U}} (s^{\prime}, \mathbf{u}^{\prime})&=  f_{\mathbf{U}}(\mathbf{u}^{\prime}) f_{S,\mathbf{U}}^*(s^{\prime}, \mathbf{u}^{\prime}; \boldsymbol{\mu}, \boldsymbol{\Sigma}) \\
	f_{\mathbf{X}^{\prime}} (\mathbf{x}^{\prime})  &= \vert s^{\prime} \vert^{1-d} f_{S^{\prime}, \mathbf{U}^{\prime}} (s^{\prime}, \mathbf{u}^{\prime})
\end{align}
where the term $f_{\mathbf{U}}(\mathbf{u}^{\prime})$ is retained because the proposed transform preserves the directional variable.

Second, for a given $\mathbf{u} = \boldsymbol{\xi}$, let $F_S(s) = F_{S|\mathbf{U}}(s | \boldsymbol{\xi})$ and $F^*_S(s) = F^*_{S|\mathbf{U}}(s | \boldsymbol{\xi})$, denote the CDFs of $f_{S|\mathbf{U}}(s | \mathbf{u})$ and its corresponding skewed term in (\ref{eq:PDF_BRFSN_FixU_Approx}). Due to our focus on 2-D case in this paper, we call this transform skew-Rayleigh-based DMPT (DMPT-SR), denoted by $T_{DMPT}^{SR}$, where $F_{S|\mathbf{U}}(s | \mathbf{u})$ is the CDF of double-sided Rayleigh distribution provided in (\ref{eq:doubleRayleighCDF}). 
Similarly, the transformed random vector, $\mathbf{X}^{\prime} = T_{DMPT}^{SR}(\mathbf{X})$,  admits the directional-radial density and Cartesian PDF, such that        
\begin{align}
	f_{S^{\prime}, \mathbf{U}^{\prime}} (s^{\prime}, \mathbf{u}^{\prime})&= f_{\mathbf{U}}(\mathbf{u}^{\prime})  f^*_{S|\mathbf{U}}(s^{\prime} | \mathbf{u}^{\prime}) \\
	f_{\mathbf{X}^{\prime}} (\mathbf{x}^{\prime})  &= \vert s^{\prime} \vert^{1-d} f_{S^{\prime}, \mathbf{U}^{\prime}} (s^{\prime}, \mathbf{u}^{\prime})
\end{align}


Essentially, the proposed transform is constructed by working with the conditional representation of the joint distribution in the directional–radial coordinate. Under \textit{Assumption \ref{ass:directionalSkewFunc}}, the skewing function restricts the support of the distribution to the line $\mathbb{L}(\mathbf{u})$, such that the joint distribution effectively reduces to a one-dimensional conditional distribution along a fixed direction. In this case, the PDF values outside the direction vanish due to $\tilde{Q}$. Therefore, the skewed PDFs in (\ref{eq: PDF_BRFSN_Approx}) and (\ref{eq:PDF_BRFSN_FixU_Approx}) can be used to construct the corresponding transforms.


\section{APPLICATION TO DECENTRALIZED SWARM NAVIGATION}

In this section, we design an algorithm to implement a decentralized swarm control law in (\ref{eq:decentralizedControlLaw}) to minimize the cost function in (\ref{eq:costFuncTotal}). For simplicity, similarly to \cite{zhu2024novel}, in this paper a binary skewing function, $Q_Y(\mathbf{x}) \in \{0, 1\}$, is applied to indicate if the location is obstacle-free, e.g., $Q_Y(\mathbf{x}) = 1$. This simplification is reasonable when the obstacle layout is known accurately.

\subsection{Association of Gaussian Components}
As shown in (\ref{eq:GMM}), the reference of the agent distribution is formulated as GMMs, a mixture model of Gaussian components, not a single MVN distribution. Therefore, to apply the DMPT directly to an individual agent, it is necessary to assign a deterministic GC $\phi_{\mathbf{X}}(\mathbf{x}|\boldsymbol{\mu}_{k}^i, \boldsymbol{\Sigma}_{k}^i)$ to every agent $\mathbf{x}_{n,k}$ at each time step. Let $L_{n,k} \in \{1, 2,\ldots, C_{k}\}$ denote a categorical random variable representing the label that the agent $\mathbf{x}_{n,k}$ is associated with the $L_{n,k}$th GCs. Then, the probability that the agent is assigned to the $i$th GC, denoted by $p_{n,k}(i) = Pr(L_{n,k} = i)$, can be calculated by 
\begin{equation}
	p_{n,k}(i) = \frac{ \omega_{k}^{i} \phi_{\mathbf{X}}(\mathbf{x}_{n,k}|\boldsymbol{\mu}_{k}^{i}, \boldsymbol{\Sigma}_{k}^{i})}{\wp_k(\mathbf{x}_{n,k})} \text{ for } i = 1, \ldots, C_{k}
	\label{eq:p_nk}
\end{equation}
which specify a Categorical distribution $L_{n,k} \sim Categorical(\mathbf{p}_{n,k})$, where $\mathbf{p}_{n,k} = [p_{n,k}(1), \ldots, p_{n,k}(C_{k})]^T$ and $\sum_{i=1}^{C_k} p_{n,k}(i) = 1$. 

For implementation purposes, the label $l_{n,k} \in \{1, 2,\ldots, C_k\}$ for corresponding agent $\mathbf{x}_{n,k}$ is kept consistent between timesteps, except for in the case of a split of a given MVN distribution, in which the agents corresponding to that distribution are randomly reassigned according to (\ref{eq:p_nk}). It is worth noting that each agent can randomly reassign itself without communications with other agents within the group.


By associating GCs with each agent, the reference for navigation is reduced from a sequence of GMMs to a sequence of individual GCs. Specifically, each agent follows a trajectory of GCs over time, instead of the full mixture model. 
Without loss of generality, we focus on a single agent denoted by $\mathbf{x}_k$. Let its associated GC trajectory be denoted by 
$\mathcal{G}_{1:T} = [\phi_{\mathbf{X},1}, \ldots, \phi_{\mathbf{X},T}]$, 
where 
$\phi_{\mathbf{X},k} = \phi_{\mathbf{X}}(\mathbf{x} \mid \boldsymbol{\mu}_k, \boldsymbol{\Sigma}_k)$ 
denotes the GC assigned to the agent at time step $k$.

\subsection{Generation of Reference locations }
Consider a set of time-varying points, $\mathcal{X}(t) = \{ \mathbf{x}_n(t)\}_{n=1}^{N} \subset \mathbb{R}^d$, evolving over time $t \in [t_0, t_1]$, where the point sets, $\mathcal{X}(t_0)$ and $\mathcal{X}(t_1)$ can be represented by two MVNs with PDFs, $\phi_{\mathbf{X}_0}(\boldsymbol{\mu}_0, \boldsymbol{\Sigma}_0)$ and $\phi_{\mathbf{X}_1}(\boldsymbol{\mu}_1, \boldsymbol{\Sigma}_1)$. Under the Wasserstein-2 metric, the optimal transport between two MVN distributions admits a closed-form affine transform denoted by $T_{Aff}$ (Lemma 2.3 in \cite{takatsu2011wasserstein}), such that
\begin{align}
	T_{Aff} : \mathbb{R}^d &\to \mathbb{R}^d \label{eq:AffineTransform} \\
	\mathbf{x} &\mapsto \mathbf{x}^{\prime} = T_{Aff}(\mathbf{x} | \phi_{\mathbf{X}_0}, \phi_{\mathbf{X}_1}) = \mathbf{A} (\mathbf{x} - \boldsymbol{\mu}_0) + \boldsymbol{\mu}_1 \nonumber
\end{align}
where
\begin{equation}
	A = \boldsymbol{\Sigma}_0^{-1/2} \left( \boldsymbol{\Sigma}_0^{1/2}  \boldsymbol{\Sigma}_1 \boldsymbol{\Sigma}_0^{1/2}   \right)^{1/2} \boldsymbol{\Sigma}_0^{-1/2}
\end{equation}
The affine transform provides a principled baseline for the MVN distribution evolution when no obstacles are presented. 

By applying the affine transform provided in (\ref{eq:AffineTransform}), first, we can transform the reference  locations for $\mathbf{x}_k$ from the $k$th step to the next step in the no-obstacle workspace, e.g., $\mathcal{W} = \mathbb{R}^2$, such that 
\begin{equation}
	\mathbf{x}_{k+1}^{Aff} = T_{Aff}(\mathbf{x} | \phi_{\mathbf{X},k}, \phi_{\mathbf{X},k+1})
	\label{eq:generateDesired_Aff}
\end{equation}    
Then, considering the obstacles in the real workspace, we generate the modified reference points by using the DMPT proposed in (\ref{eq:DMPT}), such that
\begin{equation}
	\mathbf{x}^*_{k+1} = T_{DMPT }(\mathbf{x}_{k+1}^{Aff} | \phi_{\mathbf{X},k+1}, Q_Y)
	\label{eq:referredPoint}
\end{equation}

It is noteworthy that the reference points generated in (\ref{eq:referredPoint}) are always in the obstacle-free area, e.g, $\mathbf{x}^*_{k+1} \in \mathcal{X}_{free} = \mathcal{W} \setminus \mathcal{B}$ since the binary skew function is applied. Furthermore, the simplicity of binary skew function also results in a significant decrease of computational complexity in (\ref{eq:inverseCalculation}). Specifically, the expectations of $Q_Y(S, \mathbf{u})$ in (\ref{eq: PDF_BRFSN_Approx}) and (\ref{eq:PDF_BRFSN_FixU_Approx}) are simplified to calculations of CDFs of  $\phi_S(s|0,\sigma^2_R)$ and  $f_{S|\mathbf{U}}(s|\mathbf{u})$, respectively. Due to the page limit, the implementation details are omitted.

\subsection{Decentralized Microscopic Control}

To this end, given the set of reference  points $\{\mathbf{x}^*_{n, k+1} \}_{n=1}^N$ generated according to (\ref{eq:referredPoint}), we can design the control law in (\ref{eq:decentralizedControlLaw}) for $\mathbf{x}_n(t)$, $n = 1, \ldots, N$, during $t \in (k\Delta T, (k+1)\Delta T]$ by minimizing the following weighted linear 
combination of the local artificial potential field (APF), 
\begin{equation}
	U_{LAPF}(t) = \gamma_{Att} U_{Att}(t) + \gamma_{CA} U_{CA}(t)
\end{equation}
where $\gamma_{Att}, \gamma_{CA} \in \mathbb{R}^+$ are user-defined parameters for the
attractive and the collision avoidance potentials, $U_{Att}$ and
$U_{CA}$, respectively. The attractive potential is designed to
“push” the agents toward the desired/reference locations generated in (\ref{eq:referredPoint}), such that
\begin{equation}
	U_{Att}(t) = \Vert \mathbf{x}_n (t) - \mathbf{x}_{n,k}^*\Vert^2
\end{equation}
And, the collision avoidance potential is applied to generate the repulsive force against the obstacles and other agents, which is defined by 
\begin{equation}
	U_{CA}(t) = \frac{\rho_0}{2} \left[ 1 - \frac{\rho_n (t)}{\rho_0}  \right]^2 \boldsymbol{1}[\rho_n (t) \leq \rho_0]
\end{equation}
where $\rho_n (t)$ indicates the shortest distance between the $n$th agent and the obstacles or other agents within its field of view (FOV), $\rho_0$ is a tolerance threshold. 
Finally, we can obtain the continuous control law by 
\begin{equation}
	\dot{\mathbf{x}}_{n}(t) = -\nabla U_{LAPF}(t) \text{ for } t \in  \left( k \Delta T, (k+1) \Delta T \right]
	\label{eq:continuousControlLaw}
\end{equation}

The proposed decentralized DMPT-based swarm navigation algorithm is summarized in \textbf{Algorithm} \ref{alg:dmpt_swarm}. 
\begin{algorithm}[htb]
	\caption{Decentralized DMPT-based Swarm Navigation}
	\label{alg:dmpt_swarm}
	\begin{algorithmic}[1]
		
		\Require 
		\Statex Initial agent GMM distribution: $\wp_0$
		\Statex GMM trajectory: $\mathcal{T}_{1:T} = [ \wp_1, \ldots, \wp_T ]$, 
		\Statex Obstacle map in the workspace: $Q_Y(\mathbf{x})$ for $\mathbf{x} \in \mathcal{W} \subset \mathbb{R}^2$
		\Ensure Agent trajectories $\{\mathbf{x}_{n}(t)\}$ for $n = 1,\dots,N$
		
		\For{each agent $n = 1,\dots,N$}
		\State \textbf{GC Association:} Assign GC $\phi_{X,0}$ according to the label $l_{n,0}$ sampled from the Categorical distribution specified by (\ref{eq:p_nk}) given the initial distribution $\wp_0$
		\EndFor
		
		\For{each time step $k = 0$ to $T-1$}
		\For{each agent $n = 1,\dots,N$}
		
		\State \textbf{GC Association:} Assign GC $\phi_{X,k+1}$ according to the label $l_{n,k+1}$ sampled from the Categorical distribution specified by (\ref{eq:p_nk}) given the next GMM $\wp_{k+1}$
		
		\State \textbf{Affine Transport:}
		Compute reference point without obstacles: $x^{\text{Aff}}_{n,k+1}$ according to (\ref{eq:generateDesired_Aff})
		
		\State \textbf{DMPT Correction:}
		 Apply DMPT to generate the references with obstacle: $x^*_{n,k+1}$ according to (\ref{eq:referredPoint})
		
		\State \textbf{Local Control:}
		 Compute the continuous control $\dot{\mathbf{x}}_{n}(t)$ for $t \in  \left( k \Delta T, (k+1) \Delta T \right]$ according to (\ref{eq:continuousControlLaw})
				
		\EndFor
		\EndFor
		
	\end{algorithmic}
\end{algorithm}

\section{SIMULATION AND RESULTS}
In this section, a simulation is conducted to validate the effectiveness of the proposed DMPTs, including DMPT-SN and DMPT-SR. In addition, the proposed decentralized DMPT-based swarm navigation algorithm, provided in \textbf{Algorithm} \ref{alg:dmpt_swarm}, is also implemented based on the DMPT-SN and DMPT-SR, and compared to the global APF approach on the same swarm navigation problem in \cite{gao2025swarmcvt}.  


All simulations are conducted with MATLAB R2026a code on a desktop setup with AMD Ryzen 9 9950X3D @ 5.76 GHz, 64 GB DDR5 6400 MHz RAM, and AMD Radeon RX 6950 XT @ 16 GB memory.






\subsection{Effectiveness of Directional Mass-Preserved Transforms}
\begin{figure}[htbp]
	\centering
	\begin{subfigure}[t]{0.29\columnwidth}
		\centering
		\includegraphics[height=2.70cm]{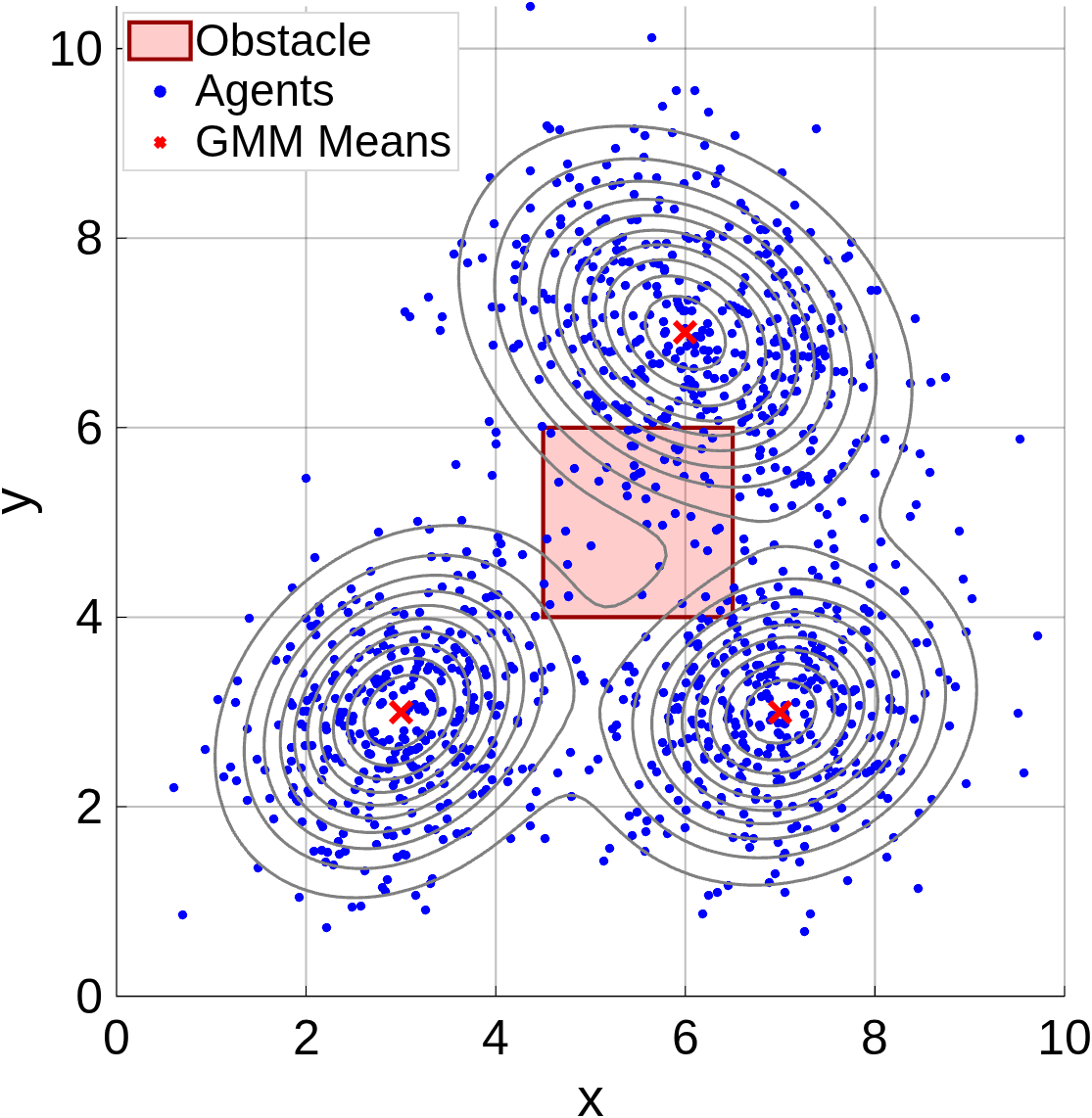}
		\caption{}
		\label{fig:original-dist}
	\end{subfigure}
	\hfill
	\begin{subfigure}[t]{0.31\columnwidth}
		\centering
		\includegraphics[height=2.70cm]{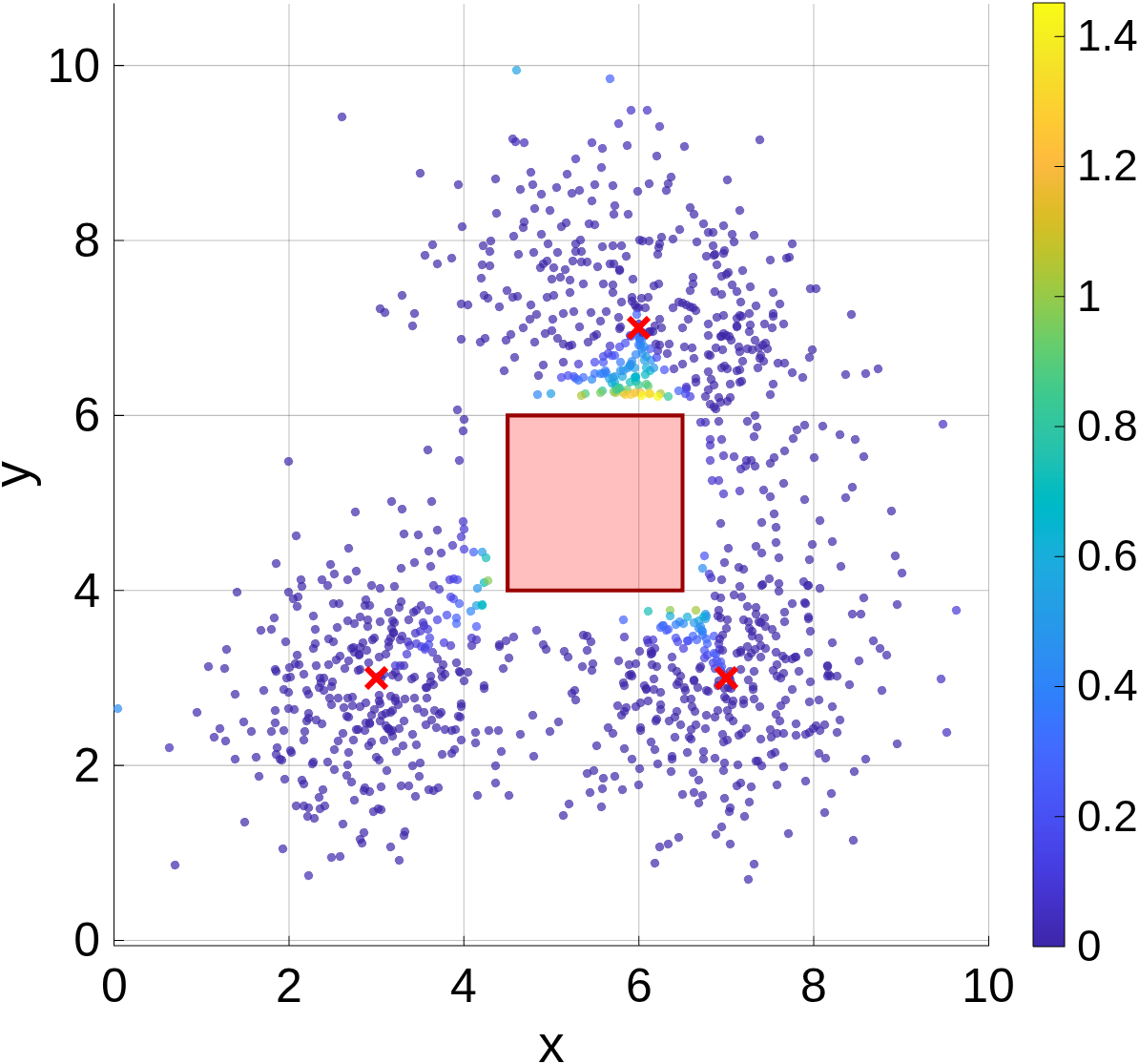}
		\caption{}
		\label{fig:gaussian-transform}
	\end{subfigure}
	\hfill
	\begin{subfigure}[t]{0.31\columnwidth}
		\centering
		\includegraphics[height=2.70cm]{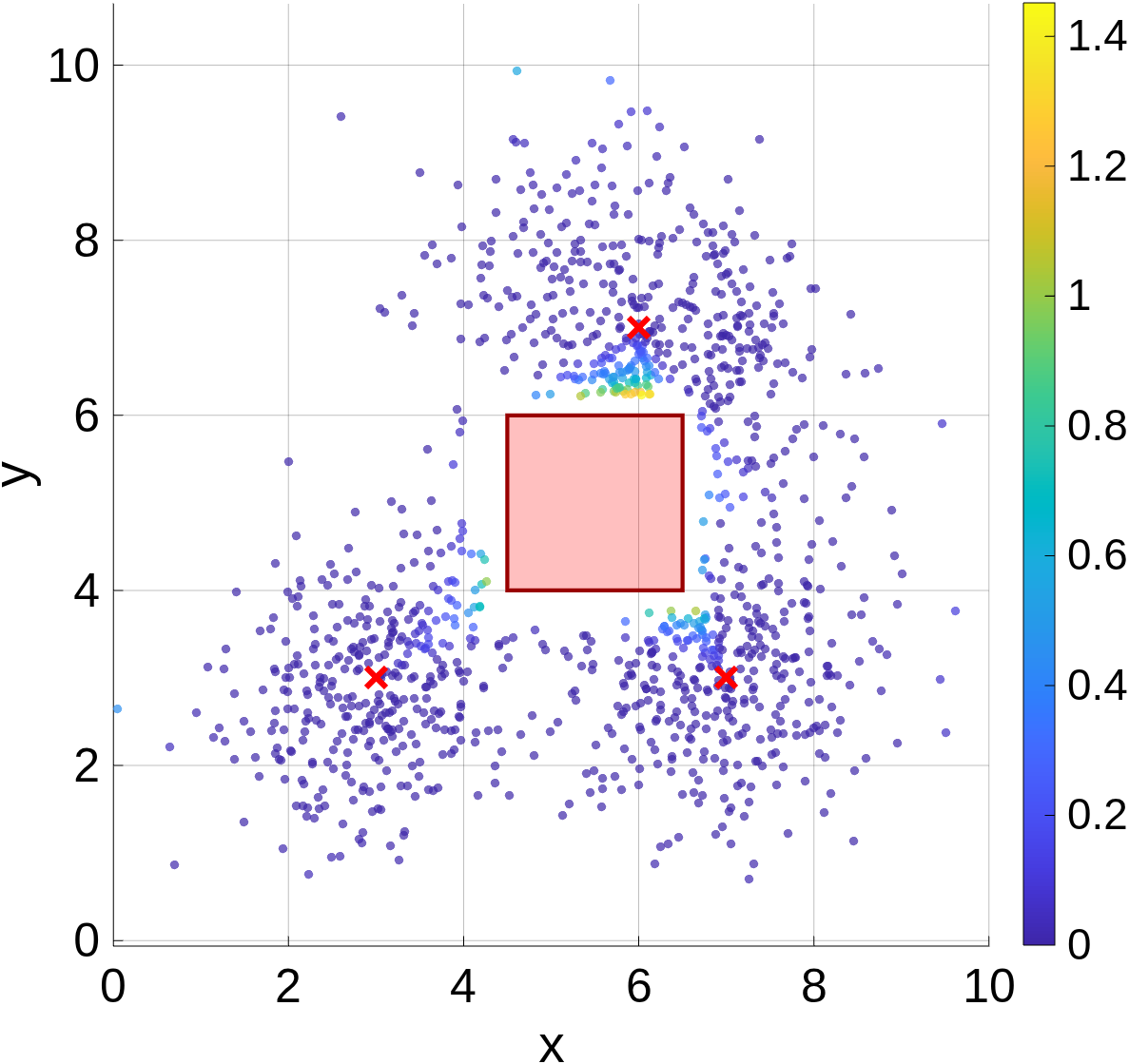}
		\caption{}
		\label{fig:rayleigh-transform}
	\end{subfigure}
	\caption{Comparison of the original GMM distribution with DMPT-SN and DMPT-SR based transports. (a) Original N=1200 Distribution; (b) DMPT-SN-based displacement magnitude; (c) DMPT-SR-based displacement magnitude.}
	\label{fig:transport-comparison}
\end{figure}

The behavior of the DMPT method is shown in Figure \ref{fig:transport-comparison}. In this simulation, N=1200 agents are initialized randomly based on the original GMM distribution shown in Figure \ref{fig:original-dist}, with some agents originally projected to be inside a square obstacle in the middle. After applying the DMPT-SN (Figure \ref{fig:gaussian-transform}) and DMPT-SR (Figure \ref{fig:rayleigh-transform}), all agents are transported out of the obstacle. 
In Figures \ref{fig:gaussian-transform} and \ref{fig:rayleigh-transform}, agents are colored by the magnitude of the distance between their initial projected position and their updated corrected positions by the DMPTs. These figures show that agents projected to be inside or too close to an obstacle are corrected by a large amount, whereas those that do not need correction are barely moved. 

\begin{figure}[htbp]
	\centering
	\includegraphics[width=\columnwidth]{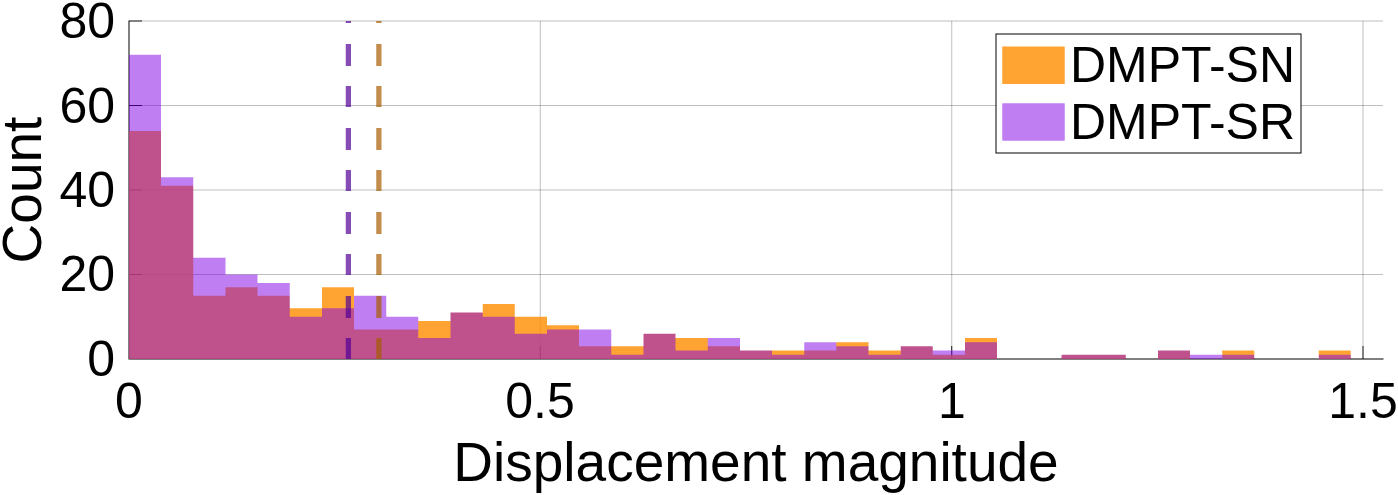}
	\caption{Histogram of difference between original and corrected agent positions.}
	\label{fig:vlsrHistogram}
\end{figure}

In addition, the performances between these two transforms are compared in Figure \ref{fig:vlsrHistogram}, where the histograms of the displacement magnitude caused by these two transforms are presented. It shows that the DMPT-SR-based method (shown in purple) moves agents by a more minimal amount compared to the DMPT-SN-based method (orange). The dotted lines represent the mean displacements for each case, with the DMPT-SR-based method showing a lower mean, which directly corresponds to lower energy costs.

We approximate the Wasserstein-2 (W2) distance between the original GMM and the transformed distributions by the following empirical approximate distance, 
\begin{equation}
	\hat{D}_{W2} = \sqrt{\frac{1}{N} \sum_{n=1}^{N} \left\| T_{DMPT}(\mathbf{x}_n) - \mathbf{x}_n \right\|^2}
	\label{transformEnergy}
\end{equation}
which is an upper bound of the W2 distance. We find that the approximated W2 disturbance is $0.3039$ km for the DMPT-SN and $0.2666$ km for the DMPT-SR, confirming that the distance and energy are reduced with the DMPT-SR-based correction.

\subsection{Simulations for Swarm Navigation}
In our previous work \cite{gao2025swarmcvt}, we proposed a Gaussian Centroidal Voronoi Tessellation (GCVT) based path planning algorithm for robotic swarms, where the trajectory of GMMs, $\mathcal{T}_{1:T} = [ \wp_1, \ldots, \wp_T ]$, is generated based on the GCVT cells, then applied a global APF-based microscopic control to all agents to follow the GMM references. Because of the utility of the global APF-based cost function, the microscopic control in \cite{gao2025swarmcvt} is a centralized method. To evaluate the performance of the proposed DMPT-based decentralized microscopic control, we replace the global APF-based microscopic control with our decentralized DMPT-based swarm control, using the same GMM trajectory reference generated from the same obstacle layout in the same swarm path-planning simulation, with the end goal of agents approximating $\wp_T$ remaining the same.

\begin{figure*}[htp]
	\centering
	
	\begin{subfigure}[t]{0.65\columnwidth}
		\centering
		\includegraphics[height=4.5cm]{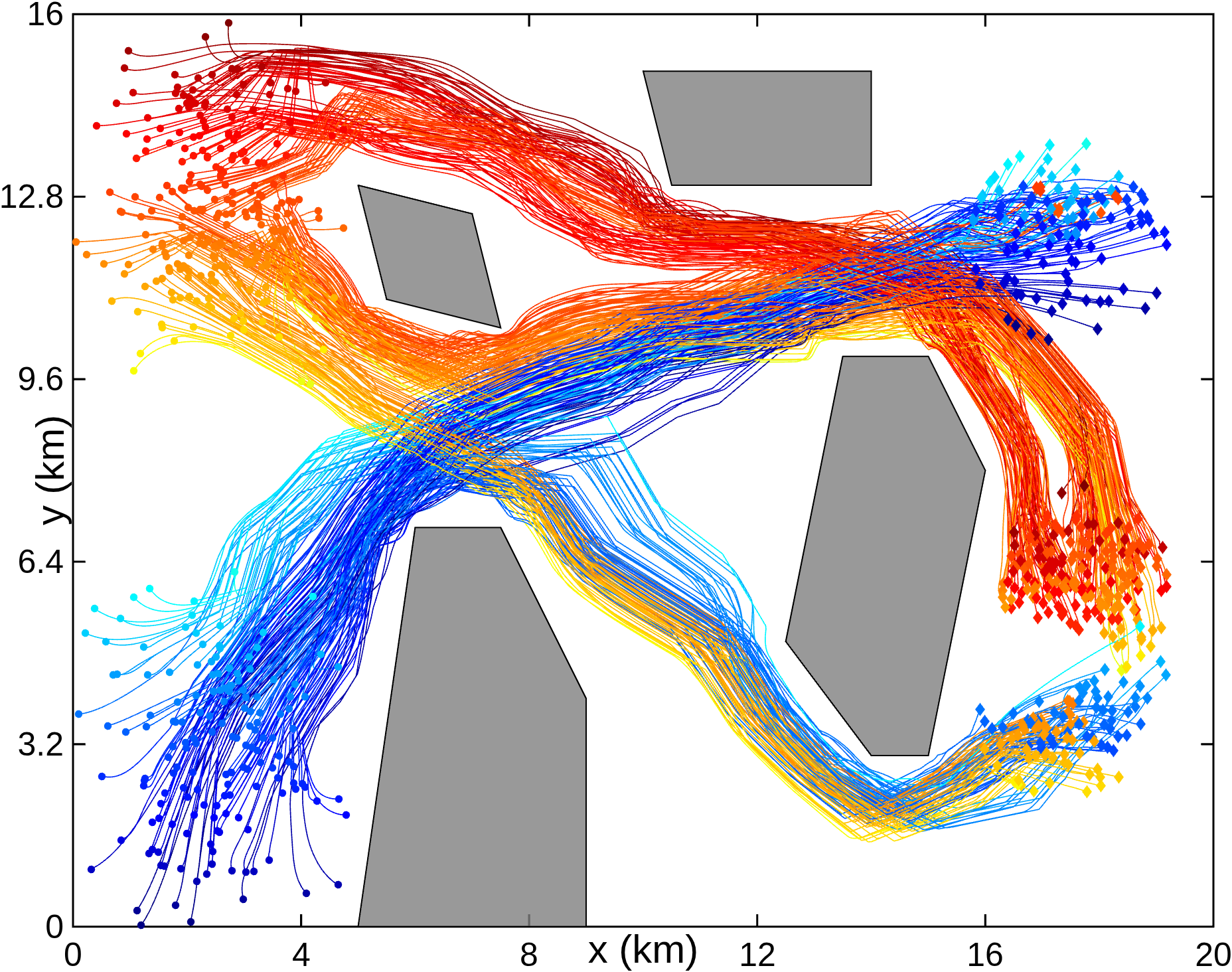}
		\caption{}
	\end{subfigure}
	\hfill
	\begin{subfigure}[t]{0.65\columnwidth}
		\centering
		\includegraphics[height=4.5cm]{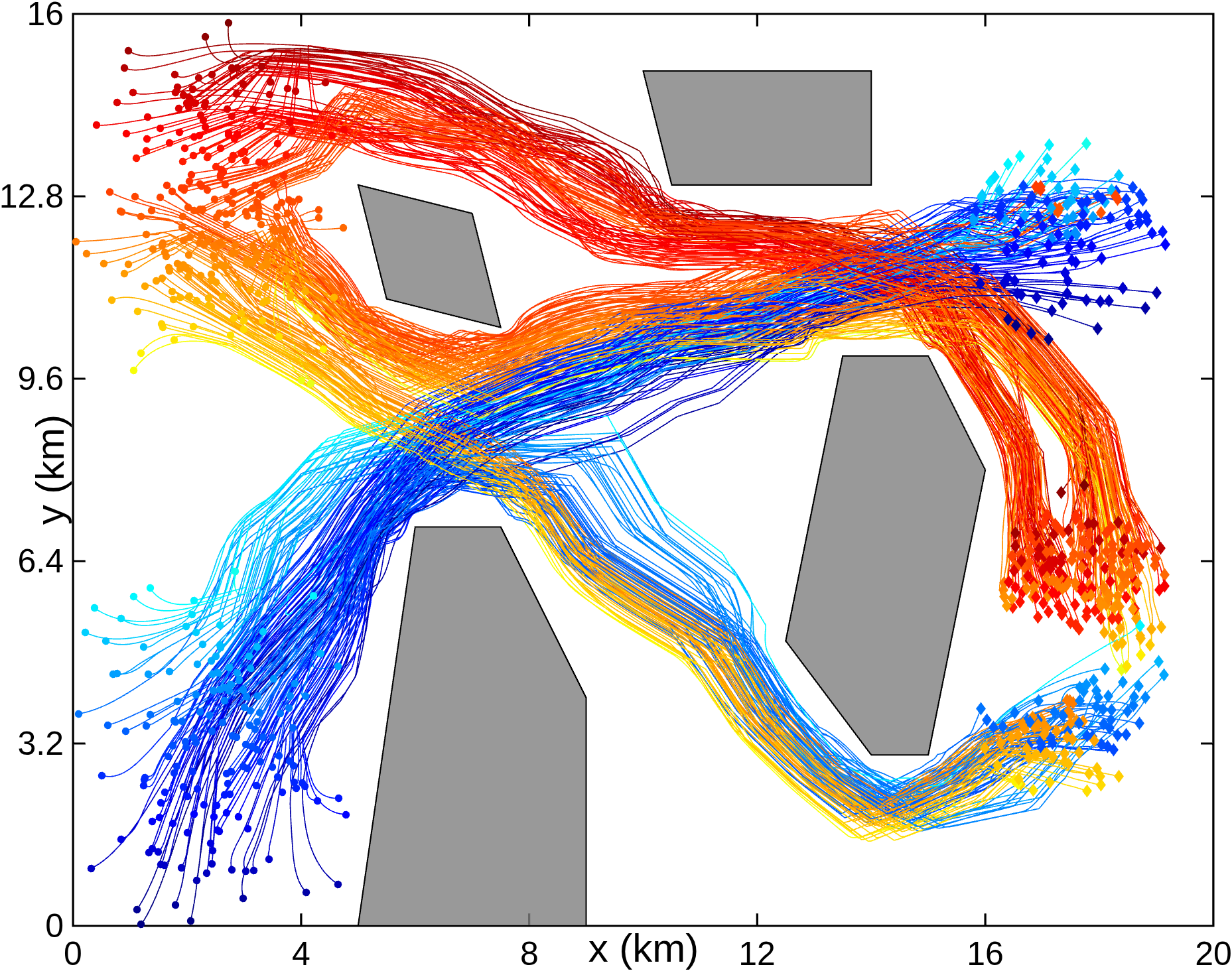}
		\caption{}
	\end{subfigure}
	\hfill
	\begin{subfigure}[t]{0.65\columnwidth}
		\centering
		\includegraphics[height=4.5cm]{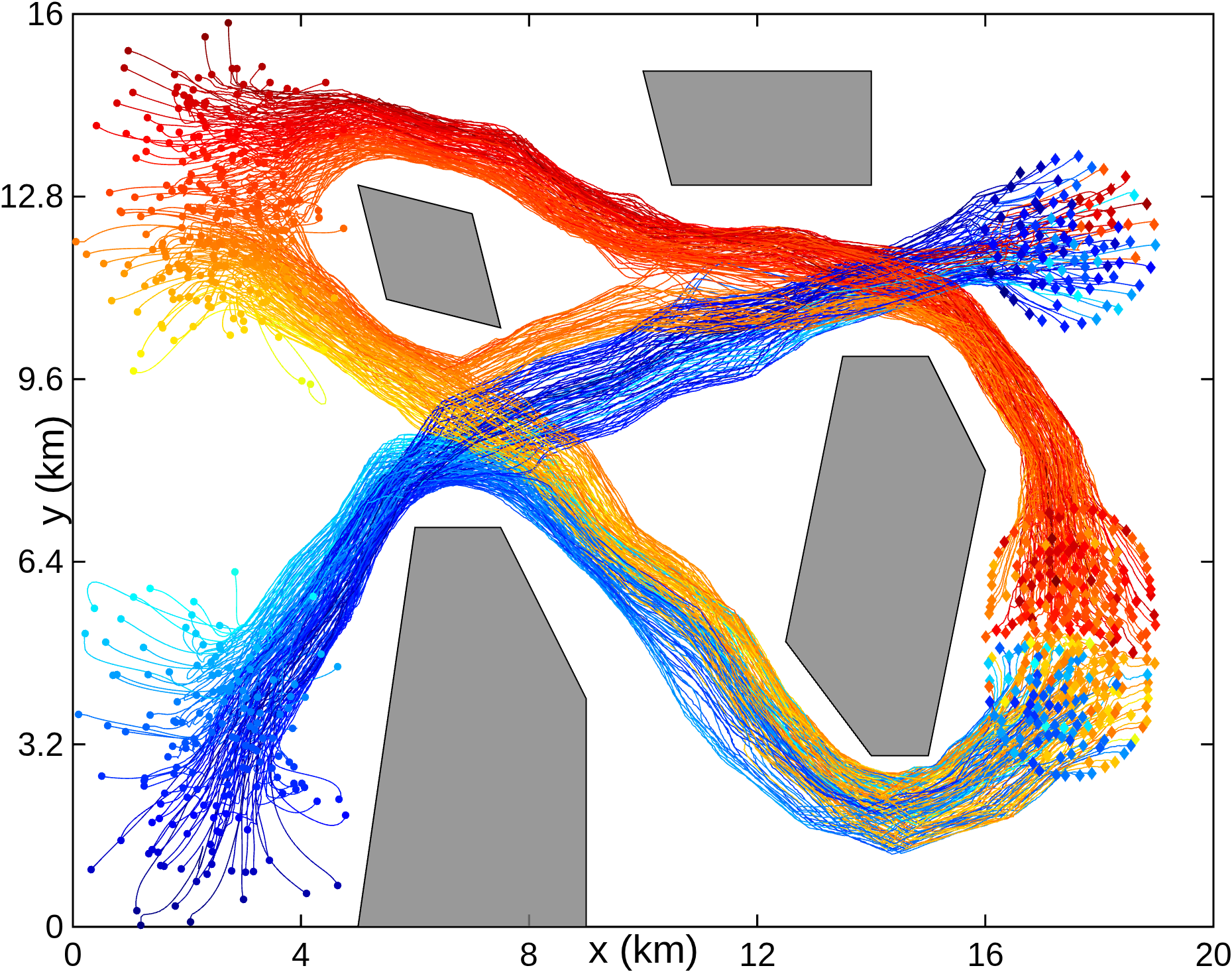}
		\caption{}
	\end{subfigure}
	\caption{Trajectories of $N=400$ agents generated by  (a) the decentralized DMPT-SN-based microscopic control, (b) the decentralized DMPT-SR-based microscopic control, and (c) the centralized global APF-based microscopic control. The trajectory begins on the left with agents generated based on a specified initial distribution, and ends by approximating a final distribution on the right.}
	\label{fig:Traj}
\end{figure*}

Figure \ref{fig:Traj} shows the single-run trajectories of $N = 400$ agents  generated from decentralized DMPT-SN, DMPT-SR, and centralized global APF-based microscopic controls, respectively. TABLE \ref{tab:performance-comparison} shows the results from running 25 trials per method with $N=400$ agents, showing significantly faster simulation time for our proposed methods compared to APF, while also showing modest improvements in distance and energy metrics. 

\begin{table}[htbp]
	\centering
	\caption{Performance Comparison for N=400}
	\label{tab:performance-comparison}
	\begin{tabular}{l|c|c|c}
		\hline
		           & $T$ (s)                    & Distance $\bar{D}$ (km)     & Energy $\bar{E}$ (J/kg)      \\
		\hline
		DMPT-SN   & $\mathbf{1.00 \pm 0.02}$   & $19.69 \pm 0.04$            & $0.9578 \pm 0.0028$          \\
		DMPT-SR   & $\mathbf{1.00 \pm 0.01}$   & $\mathbf{19.66 \pm 0.04}$   & $\mathbf{0.9559 \pm 0.0029}$ \\
		APF        & $71.29 \pm 1.10$          & $20.21 \pm 0.05$            & $1.0465 \pm 0.0044$          \\
		\hline
	\end{tabular}
\end{table}

Here, the average distance, $\bar{D}$, and the average energy, $\bar{E}$, per agent are given by:
\begin{align}
	\bar{D} &= \frac{1}{N} \sum_{k=1}^{T-1} \sum_{n=1}^{N}  \left\| \mathbf{x}_{n,k+1} - \mathbf{x}_{n,k} \right\| \\
	\bar{E} &= \frac{1}{N} \sum_{k=1}^{T-1} \sum_{n=1}^{N}  \left\| \mathbf{x}_{n,k+1} - \mathbf{x}_{n,k} \right\|^2,
	\label{energy}
\end{align}
The average energy is proportional to the real-world energy when each time step in the simulation is the same length.

\begin{table}[htbp]
	\centering
	\caption{Computation Time (s) vs.\ Number of Agents $N$}
	\label{tab:time-scaling}
	\begin{tabular}{l|c|c|c}
		\hline
		$N$ & APF & DMPT-SN & DMPT-SR \\
		\hline
		200 & $33.86 \pm 0.08$ & $\mathbf{0.71 \pm 0.01}$ & $\mathbf{0.71 \pm 0.01}$  \\
		300 & $52.06 \pm 0.23$ & $\mathbf{0.84 \pm 0.01}$ & $\mathbf{0.84 \pm 0.01}$  \\
		400 & $71.29 \pm 1.10$ & $\mathbf{1.00 \pm 0.01}$ & $\mathbf{1.00 \pm 0.01}$  \\
		500 & $96.96 \pm 2.01$ & $1.21 \pm 0.02$          & $\mathbf{1.20 \pm 0.02}$  \\
		\hline
	\end{tabular}
\end{table}

TABLE \ref{tab:time-scaling} illustrate the scaling of computational time with respect to the number of agents, $N$. The DMPT-SN- and DMPT-SR-based microscopic control methods consistently outperform the APF-based method in runtime across all tested agent counts and exhibit better scalability. 


We note that, because of the decentralized nature of our method, in a real-world application, it would scale in constant time rather than linearly. In contrast, the global APF-based microcontrol method in \cite{gao2025swarmcvt} is centralized, so the computational time roughly reflects real-world scaling.

\section{CONCLUSIONS AND FUTURE WORKS}
This paper presents a decentralized microscopic control framework for swarm navigation in obstacle environments using a directional mass-preserving transform. By combining affine transport under the Wasserstein-2 metric with the proposed DMPT, obstacle-aware reference locations are generated without inter-agent communication. A local APF-based controller then enables fully decentralized execution. The proposed framework decouples global distribution tracking into independent local control problems, significantly reducing communication and computational burden. Simulation results demonstrate improved efficiency and energy performance compared to conventional APF-based methods.

The current framework is developed under the relatively strong \textit{Assumption \ref{ass:directionalSkewFunc}}, in which the obstacle-avoidance direction of each agent is constrained to align with the direction toward the mean. This assumption improves tractability, but may limit the flexibility of the proposed transform in more complex environments. A promising direction for future research is to develop a dynamic DMPT in which the transport direction can adapt to the motion of individual agents and the local obstacle configuration. Future work will also investigate extensions to dynamic environments with online updates, theoretical guarantees on convergence and stability, and validation on real robotic platforms.


\section{APPENDIX}
\subsection{Proof of Theorem \ref{theorem:TwoDistributionsRepresentation}}
\label{proof:Theorem-1}
\begin{proof}
According to \textit{Definition \ref{def:distributionRepresentation}}, the CDFs, $F_i(s_n)$ for $i=1, 2$, satisfy the following inequalities,
\begin{align}
	F_i(s_n) &> \frac{n - 1}{N}  \nonumber\\
	1-F_i(s_n) &> \frac{N- n}{N} \text{ for } n = 1,\ldots, N 
\end{align}
Then, we have $F_i(s_n) \in (\frac{n-1}{N}, \frac{n}{N})$ for $i = 1, 2$. Finally, we can have $\vert F_1(s_n) - F_2(s_n) \vert \in [0, \frac{1}{N})$. 
\end{proof}


\bibliographystyle{IEEEtran}
\bibliography{ref_Ping_DPMT}

\end{document}